\pdfoutput=1

\documentclass[11pt]{article}

\usepackage[]{acl}

\usepackage{times}
\usepackage{latexsym}

\usepackage[T1]{fontenc}

\usepackage[utf8]{inputenc}

\usepackage{microtype}

\title{Temporal Knowledge Graph Reasoning with Low-rank \\ and Model-agnostic Representations}

\author{
  Ioannis Dikeoulias$^1$, Saadullah Amin$^2$, G\"unter Neumann$^{1,2}$\\\\
  $^1$Department of Computer Science\quad $^2$Department of Language Science and Technology\\
  Saarland Informatics Campus, D3.2, Saarland University, Saarbrücken, Germany\\
  German Research Center for Artificial Intelligence (DFKI), Saarbrücken, Germany\\
  \small\{\texttt{ioannis.dikeoulias,saadullah.amin,guenter.neumann}\}\texttt{@dfki.de}
}
  
\usepackage{subfigure}
\usepackage{mathtools}
\usepackage{amsmath}
\usepackage{amssymb}
\usepackage{amsthm}
\usepackage{bbold}
\usepackage{booktabs}
\usepackage{times}
\usepackage{latexsym}
\usepackage{amsmath}
\usepackage{amssymb}
\usepackage{float}
\usepackage{multirow}
\usepackage{graphicx}
\usepackage{listings}
\usepackage{pifont}

\newcommand{\xmark}{\ding{55}}%

\usepackage{xspace}
\usepackage[capitalise]{cleveref}
\usepackage{paralist}

\begin{document}

\maketitle

\begin{abstract}
Temporal knowledge graph completion (TKGC) has become a popular approach for reasoning over the event and temporal knowledge graphs, targeting the completion of knowledge with accurate but missing information.
In this context, tensor decomposition has successfully modeled interactions between entities and relations.
Their effectiveness in static knowledge graph completion motivates us to introduce Time-LowFER, a family of parameter-efficient and time-aware extensions of the low-rank tensor factorization model LowFER.
Noting several limitations in current approaches to represent time, we propose a cycle-aware time-encoding scheme for time features, which is model-agnostic and offers a more generalized representation of time. 
We implement our methods in a unified temporal knowledge graph embedding framework, focusing on time-sensitive data processing.
The experiments show that our proposed methods perform on par or better than the state-of-the-art semantic matching models on two benchmarks.
\end{abstract}

\section{Introduction}
\begin{figure}[!t]
    \centering
    \includegraphics[width=1.0\linewidth]{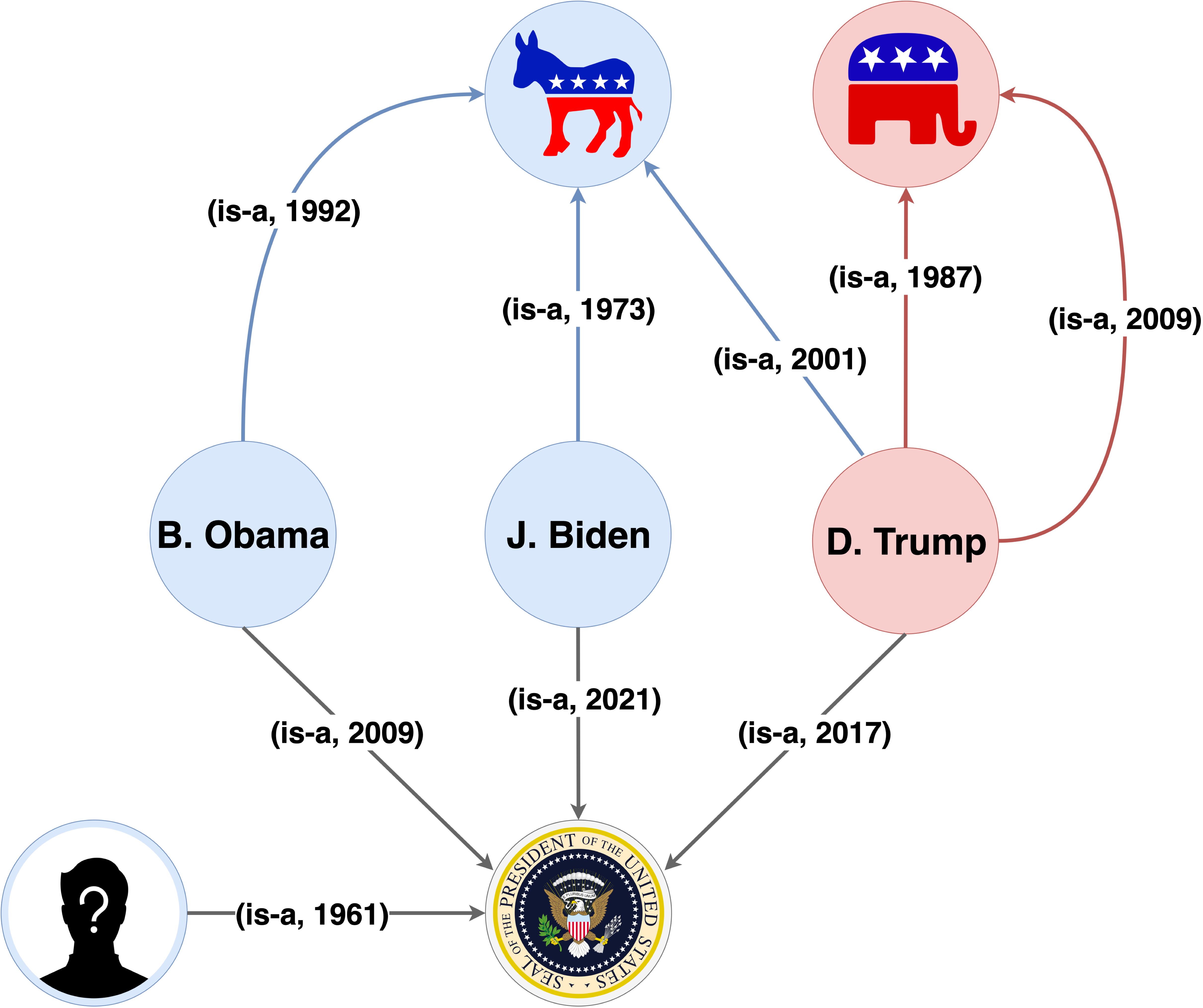}
    \caption{
    Time-sensitive relationships of U.S. presidents with the democratic and republican parties along with their timestamps. 
    Temporal knowledge graph completion (TKGC) aim to perform time-aware reasoning over the KG to answer questions of type \textit{(?, is-a, US\_president, 1961)} with an answer \textit{J. F. Kennedy}.
    }
    \label{fig:tkgc_example}
\end{figure}

Knowledge graphs offer promising technologies to structure and organize common-sense and domain-specific knowledge and form the information basis for many anticipated technological foundations.
Their importance is signified by downstream applications, including speech recognition, sentiment analysis, and knowledge base question answering \cite{dai2020survey}.
In this context, event and temporal knowledge graphs prove worthy successors of static knowledge graphs, targeting the augmentation of static relational data with temporal meta information.
\cref{fig:tkgc_example} presents a temporal sub-graph from Wikidata \cite{vrandevcic2012wikidata}, where we are interested in answering the question:
\\
\indent \emph{Who was the president of the U.S. in 1961?}
\\

Temporal information opens new opportunities for many time-sensitive domains, including time-series forecasting, biomedical event extraction, and time-sensitive crime reconstruction.
However, temporal knowledge graphs show many inconsistencies and lack data quality across different dimensions, including the accuracy, completeness, and timeliness of facts.
The quality of evolving knowledge constitutes a challenging task due to the volatile nature of knowledge.
To approach the problem of both completeness and correctness in temporal knowledge graphs, this work addresses the task of temporal link prediction and introduces time-aware extensions of the parameter efficient and expressive static embedding model LowFER \cite{amin2020lowfer}.
More precisely, we formulate the main contributions of this paper as follows:
\begin{itemize}
	\item We identify characteristic time-extension themes for extending static knowledge graph embedding models.
	\item We propose Time-LowFER, a family of time-aware and parameter efficient extensions of LowFER to temporal knowledge graphs.
	\item We identify limitations in temporal representation learning and propose a time-sensitive encoding scheme based on multi-recurrent cycle-aware time decomposition.
	\item We introduce a unified time-aware knowledge graph embedding framework focusing on time-sensitive data processing.
\end{itemize}

\section{Related Work}
\begin{figure}[t!]
    \centering
    \subfigure[][Inclusion]{
        \label{fig:ext1-a}
        \includegraphics[height=0.8in]{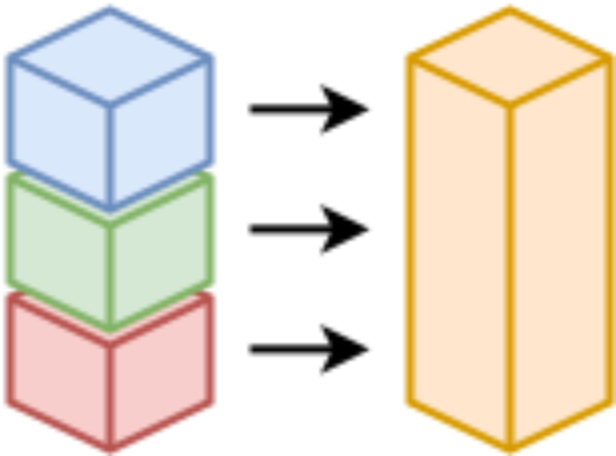}
    }
    \hspace{2em}
    \subfigure[][Modulation]{
        \label{fig:ext1-b}
        \includegraphics[height=0.8in]{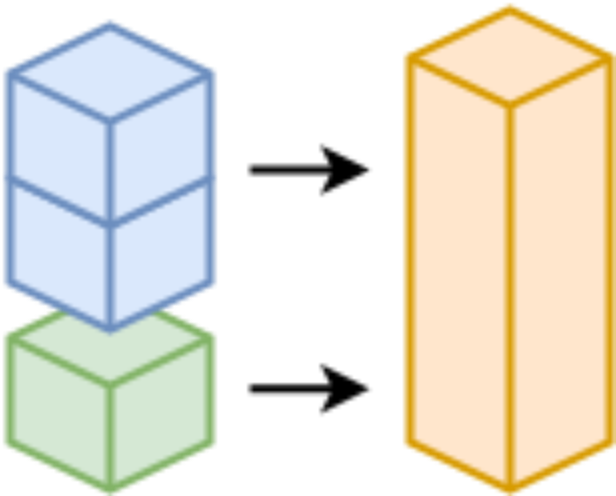}
    }\\
    \subfigure[][Regularization]{
        \label{fig:ext1-c}
        \includegraphics[height=1.0in]{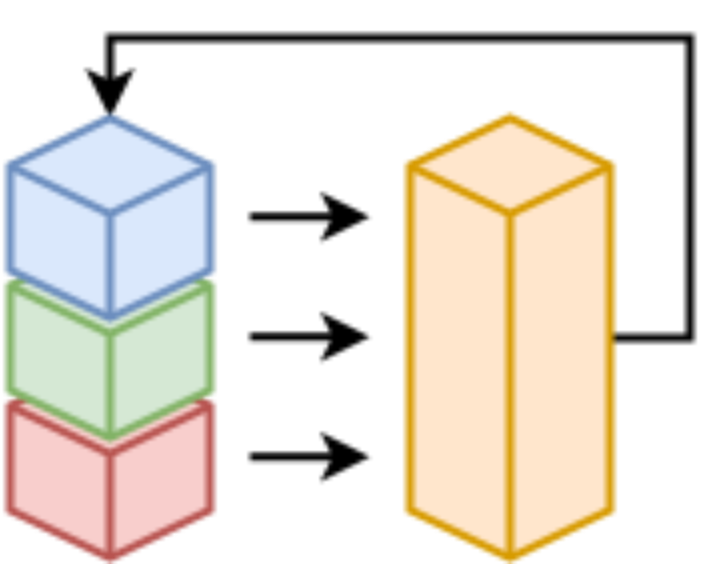}
    }
    \hspace{2em}
    \subfigure[][Aggregation]{
        \label{fig:ext1-d}
        \includegraphics[height=1.0in]{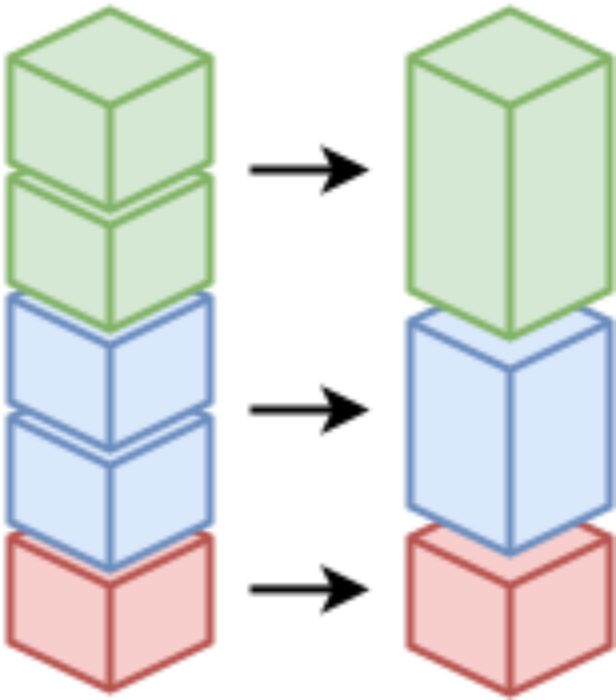}
    }
    \caption[Time extension types.]{Schematic illustration of time extension types: \subref{fig:ext1-a} inclusion-based; \subref{fig:ext1-b} modulation-based; \subref{fig:ext1-c} regularisation-based; and \subref{fig:ext1-d} aggregation-based.
    } \label{fig:time-extensions}
\end{figure}

TKGC is a prevalent task in temporal knowledge graph reasoning and targets the incompleteness and timeliness of entailed facts.
Formally, the task of TKGC is formulated as: given a temporal knowledge graph $\mathcal{G} \subseteq \mathcal{E} \times \mathcal{R} \times \mathcal{E} \times \mathcal{T}$ of quadruples $(s, p, o, t)$ where $s, o \in \mathcal{E}$ represent entities, $p \in \mathcal{P}$ represent predicates and $t \in \mathcal{T}$ represent timestamps, the task is to answer either the query $(s, p, ?, t)$ or $(?, p, o, t)$.
For the sake of completeness, we refer to $s$ and $o$ also as subject (head) and object (tail) entities, and to $p$ as relation (predicate), also commonly denoted by $r$.
TKGC approaches are divided into (a) \emph{geometric embedding models} using distance-based scoring functions, (b) \emph{semantic matching models} using similarity-based scoring functions, and (c) \emph{deep learning models}.
In this work, we focus on the group of semantic matching models, commonly referred to as factorization-based models.
The most prominent static models in this area are DistMult \cite{yang2014embedding}, SimplE \cite{kazemi2018simple}, ComplEx \cite{trouillon2016complex} and TuckER \cite{balavzevic2019tucker}.
Further, noting characteristic patterns throughout several time-aware extensions of static embedding models, we identified four distinct temporal extension themes
\cref{fig:ext1-b}: (1) inclusion-based, (2) feature-based, (3) regularization-based, and (4) aggregation-based extensions.

\subsection{Inclusion-based extensions}
Inclusion-based approaches represent extensions, where time features are exposed directly to the underlying embedding model.
Time features are considered individual input signals, favoring a more expressive inclusion of time information within the model.
A prominent example for inclusion-based time extension is TTransE \cite{leblay2018deriving}, representing time as temporal translation of entity-relation features.
Similarly, TeRo \cite{xu2020tero} considers individual time features via temporal rotation of entity features.
Both models incorporate time as a separate feature and can learn more expressive interactions between input features.

\subsection{Modulation-based extensions}
Modulation-based extensions are the most commonly used approaches for the time-aware extension of static embedding models.
In this context, modulation describes regulating a base signal using a separate modulation signal, i.e., time-based relation modulation allows for a time-sensitive parametrization of relation embeddings.
Most common approaches are TNTComplEx \cite{lacroix2020tensor} and TuckERTNT \cite{shao2021tucker}, which extend their static base models ComplEx \cite{trouillon2016complex} and TuckER \cite{balavzevic2019tucker} via a temporal modulation of either entity or relation features.
Similarly, diachronic embeddings \cite{goel2020diachronic} represent a model-agnostic and time-aware extension of static embedding models via modulation of entity-specific parameters.
In contrast to inclusion-based extensions, modulation-based extensions do not expose time directly to the underlying embedding model.
Therefore, modulation is model-agnostic and similarly allows for a parameter-efficient extension of static embedding models.

\subsection{Regularization-based extensions}
Regularization-based extensions include techniques that impose consistency constraints on learnable feature representations.
A commonly used scheme for time-aware regularization is temporal smoothness regularizers, which leverage the semantic nearness of the nearby timestamps.
Both TuckERTNT \cite{shao2021tucker}, and TNTComplEx \cite{lacroix2020tensor} implement this type of smoothness regularization scheme.
The authors minimize the nuclear $p$-norm (N3, 3-norm) of the discrete derivative of two nearby time embeddings, effectively penalizing sharp time gradients.
Likewise, for TransE-TAE \cite{jiang2016towards} the authors introduce time-wise regularization schemes that enforce constraints for the temporal ordering or disjointness of facts.
This way, embedding models are less prone to overfitting and allow for improved generalization to underlying data.

\subsection{Aggregation-based extensions}
Aggregation-based extensions leverage the compositionality of features such as clustering, grouping, averaging, and sampling.
An application of this idea has been implemented by TeMP \cite{wu2020temp}, which introduces time-based subgraph clusters that group information together that occurs within a specific time range.
Similarly, ATiSE \cite{xu2020temporal} and TeRo \cite{xu2020tero} introduce a temporal granularity parameter that varies the temporal sampling rate at which facts are discretized over time.

\section{Low-rank Representation}

Tensor factorization models decompose the order-3 and order-4 binary tensor into a compressed tensor and a set of factor matrices, respectively, for static and temporal KGs.
TuckER \cite{balavzevic2019tucker} proposed a Tucker decomposition model for static KGs and showed that existing semantic matching models could be subsumed in their formulation. 
Noting the cubic growth of core tensor in TuckER, LowFER \cite{amin2020lowfer} proposed an efficient parameter initialization of the core tensor using low-rank factorized bilinear pooling.
Due to its ability to handle arbitrary relations (\emph{fully expressive}), parameter efficiency, generalization abilities, and state-of-the-art performance in embedding-based models for static KGs \cite{zhu2021neural}, we extend it to the temporal KGs.

\subsection{LowFER}

LowFER \cite{amin2020lowfer} introduces a low-rank decomposition of the core tensor in TuckER, reducing the parameter growth from $\mathcal{O}(d^3)$ to $\mathcal{O}(kd^2)$, with $d$ and $k$ being the embedding dimension and factorization rank, respectively.
Given subject entity and relation  embeddings $\mathbf{e}_s$ and $\mathbf{e}_p$, LowFER approximates the interaction tensor using two low-rank projection matrices $\mathbf{U} \in \mathbb{R}^{d_e\times kd_e}$ and $\mathbf{V} \in \mathbb{R}^{d_r\times kd_e}$.
More specifically, both entity and relation features are projected to high-dimensional spaces $\mathbf{U}^T\mathbf{e}_s$ and $\mathbf{V}^T\mathbf{e}_p$, followed by Hadamard product (denoted by $\circ$) and $k$-sized non-overlapping summation pooling:
\begin{align*}
    f(e_s, e_p, e_o) = \langle \mathbf{g}(e_s, e_p), \mathbf{e}_o \rangle
\end{align*}
where $\langle.,.\rangle$ defines dot product and $\mathbf{g}$ represents a vector-valued function, performing factorized bilinear pooling:
\begin{align*}
    g(e_s, e_p) := \text{SumPool}(\mathbf{U}^{T}\mathbf{e}_s \circ \mathbf{V}^{T}\mathbf{e}_p, k),
\end{align*}
and $\mathbf{e}_o$ represents the target entity.
LowFER can generalize the TuckER model.
Moreover, for $k\leq\text{min}(d_e, d_r)$ with $d_e, d_r$ as entity and relation dimensions, respectively,  LowFER is able to accurately represent TuckER's core tensor $\mathcal{W}_{\text{tucker}}$.
In addition, given the subsumption of TuckER, LowFER is equally fully expressive and thus able to represent arbitrary relations, e.g., symmetric, reflexive, and transitive, among others. 

\subsection{Time-LowFER}
This section introduces Time-LowFER, a family of time-aware extensions of the bilinear embedding model LowFER.
These include (i) LowFER-TNT: a modulation-based extension following time-relation modulation \cite{lacroix2020tensor}, (ii) LowFER-CFB: an inclusion-based extension using chained bilinear pooling, and (iii) LowFER-FTP: a reduced variant of the latter, with factorized trilinear pooling.

\subsection{Factorized Bilinear Pooling} \label{subsec:fbp}
Following existing works of \citet{lacroix2020tensor,shao2021tucker}, we propose two variants of time modulation also referred to as time modulation (T) and time-no-time modulation (TNT), which extends the factorized bilinear pooling of LowFER.
The first variant (T) performs a simple temporal modulation of relation features $e_r$ using time features $e_t$.
We refer to this extension as LowFER-T and formulate its scoring function as follows:
\begin{align*}
    f(e_s, e_p, e_o, e_t) = \langle\mathbf{g}(e_s, e_p, e_t), \mathbf{e}_o\rangle
\end{align*}
where $e_t$ is time embedding and $\mathbf{g}$ is defined as:
\begin{align*}
    g(e_s, e_p, e_t) = \mathbf{U}^T\mathbf{e}_s \circ \mathbf{V}^T(\mathbf{e}_p \odot \mathbf{e}_t)
\end{align*}
Time modulation (T), denoted by $\odot$, enables LowFER-T to learn joint time-aware representations of entity and relation features such that learned feature-to-feature interactions now incorporate the dynamics of the overlying knowledge graph with more precise predictions.
However, not all predicates are similarly affected by time or show reduced sensitivity to temporal changes in related facts.
For instance, the predicate \textit{born\_in} is not changing over time, however the predicate \textit{works\_at} (most probably) will.
To capture both dynamic and static characteristics of temporal relations, we follow \citet{lacroix2020tensor} and propose a time-no-time (TNT) variant of LowFER, which calculates a combined representation of static and dynamic relations.
We refer to this extension as LowFER-TNT and formulate the function $\mathbf{g}$ as:
\begin{align*}
    g(e_s, e_p, e_t) = \mathbf{U}^T\mathbf{e}_s \circ \mathbf{V}^T(\mathbf{e}_p^t \odot \mathbf{e}_t + \mathbf{e}_p)
\end{align*}
$e_p^t$ denotes the time-aware relation embedding and $e_p$ the static relation feature.
Both T and TNT variants of LowFER do not modify the assumptions of the underlying methods, i.e., the approximation of TuckER's core interaction tensor.
Therefore, both modulation-based extensions are fully expressive and can be seen as (time-aware) generalizations of LowFER, while equally, TNT subsumes T.

\subsection{Chained Factorized Bilinear Pooling}
Time features encode latent dynamics of evolving knowledge graphs and allow for a time-aware classification of relational links, i.e., time features set crucial constraints on feature interactions, similarly disqualifying the existence of specific graph structures for a given time range.
For instance, after \textit{(U.S.) presidential elections}, the link \textit{is\_president} will remain static for at least four years.
Similarly, the \textit{signing} of international climate agreements would imply a change of \textit{demands} for government and industry.
However, bilinear models (\cref{subsec:fbp}) are solely defined over two variables, making them less suitable for multivariate analysis, e.g., entity, relation, and time features.
We propose a multilinear method based on bilinear chaining for use in multivariate learning to overcome these limitations.

More precisely, we define a $k$-fold chaining of bilinear models through the nesting of bilinear transformations:
\begin{align*}
    & B_{k}^{C}(x_1, \dots, x_{k+1}) \\
    &= (B_{1} \circ \dots \circ B_{k})(x_1, \dots, x_{k+1})\\
    &= B_{k}(B_{k-1}(\dots, x_{k}), x_{k+1}),
\end{align*}
where $B_{1}, \dots, B_{k}$ denote bilinear transformations, $(x_1, \dots, x_{k+1})$ denote the input features and the $\circ$ here represents the function composition operator. 
Following LowFER, we introduce a time-aware extension of LowFER based on two-fold chaining of factorized bilinear methods.
\begin{align*}
    g(e_s, e_p, e_t) := \text{SumPool}((B_1 \circ B_2)(e_s, e_p, e_t), k)
\end{align*}
where $B_1$ and $B_2$ denote two nested bilinear transformations.
Similarly, the above equation can be re-written in terms of three low-rank projection matrices, $\mathbf{U}, \mathbf{V}$ as similar to LowFER and $\mathbf{Q} \in \mathbb{R}^{dr \times kd_e}$ as:
\begin{align*}
    & g(e_s, e_p, e_t) \\
    &= \text{SumPool}(\mathbf{U}^T\mathbf{e}_s \circ \mathbf{R}^T (\mathbf{V}^T\mathbf{e}_p \circ \mathbf{Q}^T\mathbf{e}_t ), k)
\end{align*}
We refer to this method as chained factorized bilinear (CFB) pooling.
CFB learns two joint representations between relation and time features and once between entities and the joint representation of time-relation features.
However, we note that the intermediate projection matrix $\mathbf{R} \in \mathbb{R}^{kd_e \times kd_e}$ is likely to share redundant parameters with both relation and time projection matrices $\mathbf{V}$ and $\mathbf{Q}$.

\subsection{Factorized Trilinear Pooling}
CFB enables the computation of fine-grained interactions between different feature spaces.
However, CFB introduces redundant parameters via the intermediate projection of joint time-relation features.
To retain the efficiency of the original low-rank bilinear method, we introduce a (reduced) specialization of the CFB, which omits the intermediate feature projection.
Replacing $\mathbf{R}$ by an identity matrix $\mathbf{I}$, we formulate the respective method via a three-way Hadamard product or entity, relation, and time features, followed by summation pooling:
\begin{align*}
    & g(e_s, e_p, e_t) \\ 
    &= \text{SumPool}(\mathbf{U}^T\mathbf{e}_s \circ \mathbf{V}^T\mathbf{e}_p \circ \mathbf{Q}^T\mathbf{e}_t, 1),
\end{align*}
Where $\mathbf{U}$, $\mathbf{V}$, and $\mathbf{Q}$ are the low-rank projection matrices, note the factorization rank, $k$ is set to 1.

\section{Cycle-aware Time Embedding}
In this section, we propose a novel extension technique for embedding time features ($\mathbf{e}_t$), which relies on multi-recurrent cycle-aware (MRCA) time decomposition and is model-agnostic.
We first explain the concept of \textit{multi-recurrence} and then show its application for \textit{cycle-aware encoding} of time features.
Temporal recurrence denotes the concept of expressing time as recurrent component within a certain time frame e.g.,  a \textit{week} occurs approximately four times per \textit{month} or equally a \textit{year} contains four \textit{seasons}.
In this context, we speak of \textit{multi-recurrence}, if a time frame is expressed in terms of multiple cycles, e.g., a \textit{year} entails four \textit{seasons}, 12 \textit{months}, \textit{52} weeks and \textit{365} days.
Multi-recurrence is expressible for all time concepts with one or more underlying recurrent cycles.
Even more, (long-term) cycles themselves are also expressible in terms of more fine-grained (short-term) cycles.
Our multi-recurrent cycle-aware (MRCA) encoding uses a mapping $\psi^{cyc}$ of timestamps $T$ to a set of recurrence encodings $C^{m \times l}$:
\begin{align*}
& \psi^{cyc}: T \rightarrow C^{m \times l}\\
& t_{i} \mapsto (\phi_{W_1}^{rec}(t_{i}), \dots, \phi_{W_m}^{rec}(t_{i})), \ 1 \leq i \leq n,
\end{align*}
where each recurrence encoding $\phi_{W_k}^{rec}$ defined upon cycle window $W_k$, uses a mapping of timestamps $T$ to a set of cycle indices $C^l$:
\begin{align*}
    & \phi_{W_k}^{rec}: T \rightarrow C^{l}\\
    & t_{i} \mapsto (c_1^{k}, \dots, c_{l}^{k}), \ 1 \leq k \leq m, 
\end{align*}
and each time component $c_{p}^{k}$ with $1 \leq p \leq l$ is defined as:
\begin{align*}
c_{p}^{k} = \begin{cases} 
      v & \text{if \textit{p-th} subcycle exists in $W_k$}\\
      0 & \text{else}
   \end{cases}.
\end{align*}
Here, $n$ denotes the number of timestamps, $m$ is the number of time windows, and $l$ is the number of recurrent subcycles.
Now, to generate the cycle-aware multi-recurrent time embedding, we consider the following five components and their respective cycle decompositions\footnote{The generation of recurrent time cycles rely on prior knowledge of empirically defined periods, e.g., a week has seven days, a month has 30 days, a year consists of approximately 365 days, e.t.c.}:
\begin{figure}[!t]
    \centering
    \includegraphics[width=1.0\linewidth]{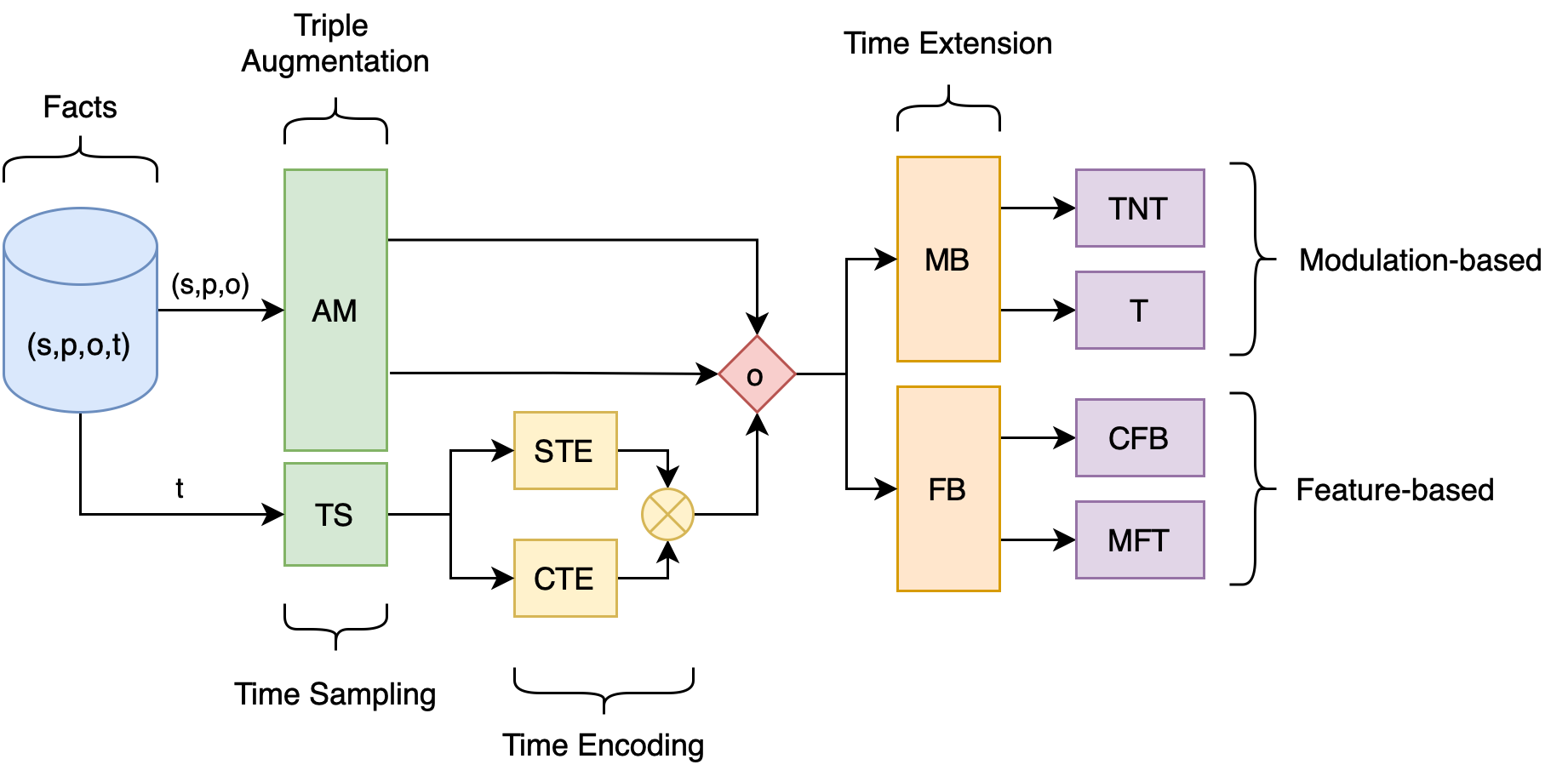}
    \caption{Overview of data processing and modeling pipeline in the \textsc{ChronoKGE} framework. A quadruple fact is processed using triple augmentation and time sampling (\emph{green}), and afterward, timestamps are decomposed into subcomponents (\emph{yellow}). Then, the processed quadruple is passed to the respective temporal extension (\emph{orange}) and model variation (\emph{violet}).}
    \label{fig:pipeline}
\end{figure}
\begin{align*}
	 &\phi_{W}^{rec}(t_{i}) &\rightarrow \ &(c_{d}^{W})\\
	 &\phi_{M}^{rec}(t_{i}) &\rightarrow \ &(c_{d}^{M}, c_{w}^{M})\\
	 &\phi_{S}^{rec}(t_{i}) &\rightarrow \ &(c_{d}^{S}, c_{w}^{S}, c_{m}^{S})\\
	 &\phi_{Y}^{rec}(t_{i}) &\rightarrow \ &(c_{d}^{Y}, c_{w}^{Y}, c_{m}^{Y}, c_{s}^{Y})\\
	 &\phi_{G}^{rec}(t_{i}) &\rightarrow \ &(c_{1}^{G}, c_{10}^{G}, c_{100}^{G}, c_{1000}^{G})\\
\end{align*}
Where $W, M, S, Y, G$ denote the weekly, monthly, seasonal, yearly, and global components, respectively, and $d, w, m, s, y$ denote the daily, weekly, monthly, seasonal, and yearly subcycles.
The components $c_{1},c_{10}, c_{100}, c_{1000}$ denote the positional year component for single years, decades, centuries and milleniums, respectively.
The final cycle-aware time encoding $\psi^{MRCA}$ is then generated by summing over all cycle decompositions of each separate recurrence mapping $\phi_{W_k}^{rec}$:
\begin{align*}
\mathbf{e}_t = \psi^{MRCA}(t) & = \sum_{i=1}^{l_1} (\phi_{G}^{rec}(t))_i + \sum_{i=1}^{l_2} (\phi_{Y}^{rec}(t))_i\\
& + \sum_{i=1}^{l_3} (\phi_{S}^{rec}(t))_i + \sum_{i=1}^{l_4} (\phi_{M}^{rec}(t))_i\\
& + \sum_{i=1}^{l_5} (\phi_{W}^{rec}(t))_i
\end{align*}
\begin{table*}[!t]
    \centering
    \begin{tabular}{lcccccccc}
        \toprule
        & \textbf{\#E} & \textbf{\#R} & \textbf{\#T} & \textbf{time span} & \textbf{gran.} & \textbf{MD.} & \textbf{TQ.} & \textbf{MC.} \\
        \midrule
        \multicolumn{1}{c}{\textbf{ICEWS14}} & 7,129 & 230 & 365 & 2014 & daily & \xmark & \xmark & \xmark \\
        \multicolumn{1}{c}{\textbf{ICEWS05-15}} & 10,488 & 251 & 4,017 & 2005-2015 & daily & \xmark & \xmark & \xmark \\
        \bottomrule
    \end{tabular}
    \caption{Data statistics and properties: No. entities (E), no. relations (R), no. timestamps (T), time span, time granularity, multiple domains (MD), temporal qualifier (TQ), and manual curation (MC).}
    \label{tab:data}
\end{table*}

\section{Experiments and Results}

\subsection{Data}

ICEWS (Integrated Crisis Early Warning System) was founded in 2008 as a DARPA program and is currently maintained by Lockheed Martin.
The conflict warning system collects news about political events from different digital and social media platforms and stores the extracted information in the associated ICEWS database.
\textsc{Icews14} is a subset of the ICEWS database, including facts from the year 2014.
It consists of 7,129 distinct entities, 230 relations, and 365 timestamps with 24 hours (daily) temporal granularity. 
\textsc{Icews05-15} is another subset of the ICEWS database, including facts between the start of 2005 and the end of 2015.
It consists of 10,488 distinct entities, 251 relations, and 4,017 timestamps with a temporal granularity of 24 hours (daily).
In \cref{tab:data} we provide an overview of the datasets.
\subsection{Implementation}
For implementation, we developed an extensible temporal knowledge representation learning framework \textsc{ChronoKGE}\footnote{\url{https://github.com/iodike/ChronoKGE}}. \cref{fig:pipeline} shows the data and modeling pipeline in the framework. We present more details in Appendix A.1. 
For training, we use the Adam \cite{kingma2014adam} optimizer with a learning rate of 0.01 and a decay rate of 0.99.
We perform 1-N scoring with binary cross-entropy loss and choose a batch size of 1000.
We further apply a label smoothing of 0.01 to the target labels.
The embedding dimension of entities, relations, and time is set to 300.
We use $k=32$ in LowFER-TNT and LowFER-CFB, and use dropout following \citet{amin2020lowfer}. 

\subsection{Baselines}
We choose the static LowFER model for our experiments to apply to our proposed temporal extensions.
Since LowFER is a semantic matching linear model, we only compare our results with the extensions of such linear models.
In particular, we use as baselines the time-aware extension of both ComplEx (TComplEx, TNTComplEx) \cite{lacroix2020tensor}, TuckER (TuckERT, TuckERTNT) \cite{shao2021tucker} as well as SimplE (DE-SimplE) and DistMult (DE-DistMult) \cite{goel2020diachronic}.

\subsection{Time and No-Time Modulation}
\begin{figure}[t!]
    \centering
    \subfigure[][ICEWS14]{
    \label{fig:time-rel-hm-a}
    \includegraphics[height=1.8in]{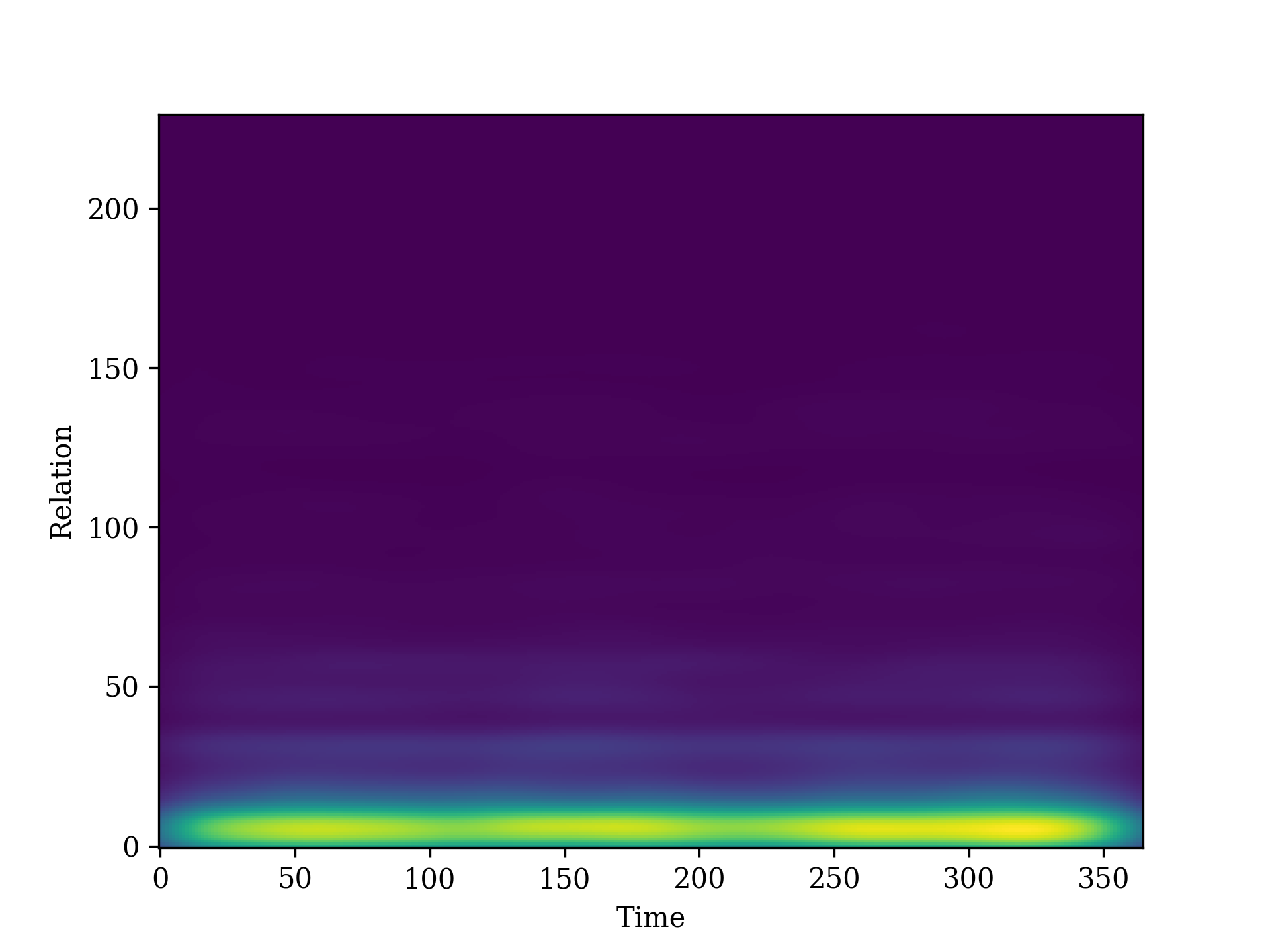}}
    \hspace{0pt}
    \subfigure[][ICEWS05-15]{
    \label{fig:time-rel-hm-b}
    \includegraphics[height=1.8in]{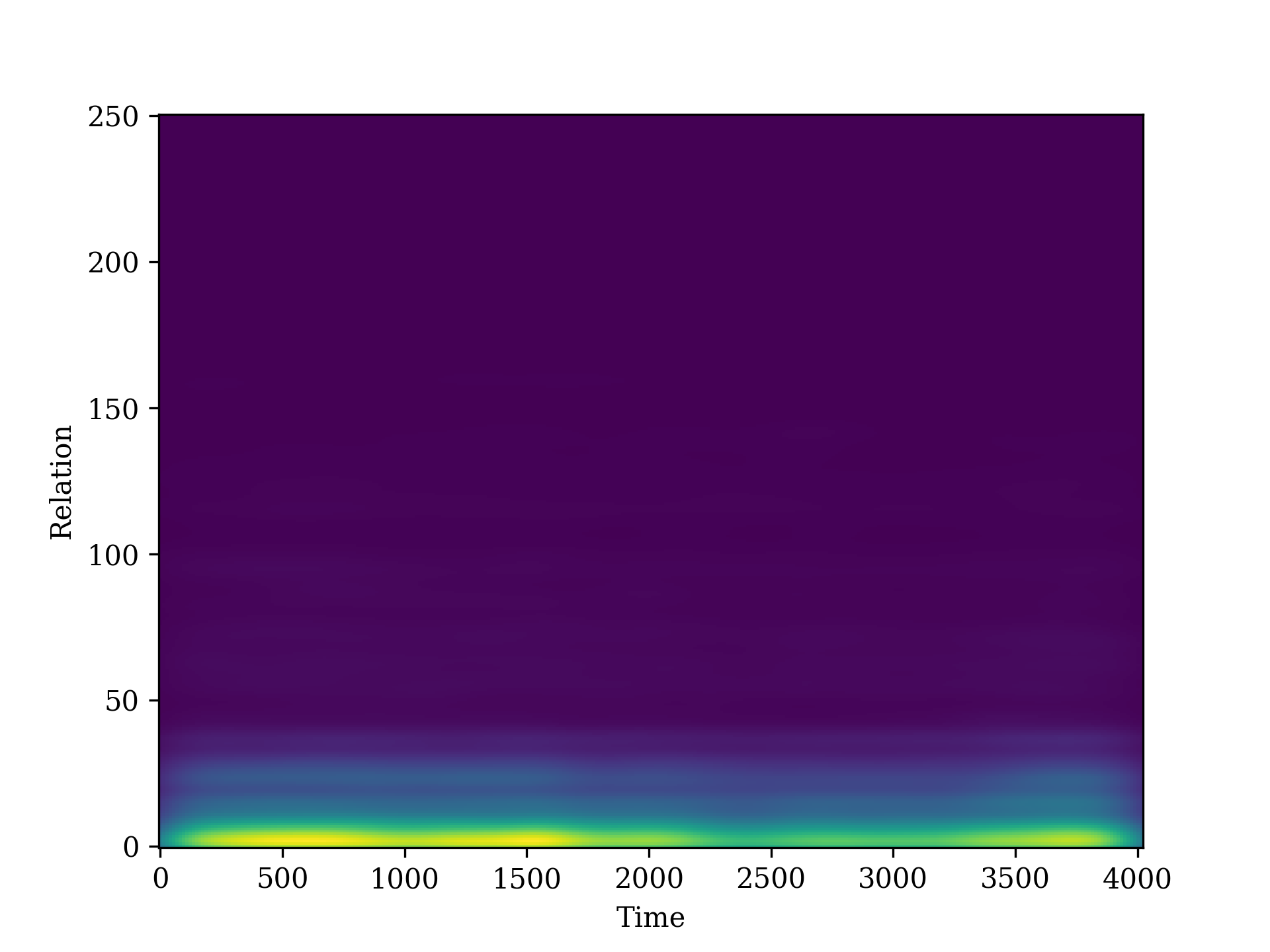}} \\
    \caption[Time-relation heatmap.]{Time-relation heatmap:
    \subref{fig:time-rel-hm-a} for ICEWS14; and
    \subref{fig:time-rel-hm-b} for ICEWS05-15.}
    \label{fig:time-rel-hm}
\end{figure}

Following the findings of \citet{shao2021tucker} and \citet{lacroix2020tensor}, we only modulate relations with time information (instead of entities).
For both ICEWS datasets, as depicted in Table \ref{tab:results-full}, we see persistent improvements in the TNT-extension over the T-extension.
It shall be noted that both TuckERTNT and TNTComplEx use time and embedding regularization schemes, where Time-LowFER extensions are reported without any regularization that can further improve the results.
While the T-extension learns a time-aware relation that primarily relies on temporal information, the TNT extends it by learning an additional (static) relation embedding as shown in \cref{fig:time-rel-hm}.
This effect is beneficial for highly frequent relations (light areas), which should not rely too strictly on time.
In contrast, temporal information is much more valuable for relations that occur less frequently (dark areas) and should be incorporated with a higher weighting.
\begin{table*}[!t]
    \centering
    \begin{tabular}{rcccccccc}
        \toprule
        & \multicolumn{4}{c}{\textsc{Icews14}}  & \multicolumn{4}{c}{\textsc{Icews05-15}} \\
        \cmidrule{2-9}
        & MRR & H@10 & H@3 & H@1 & MRR & H@10 & H@3 & H@1 \\
        \midrule
        DE-DistMult$^\diamond$ & 0.501 & 0.708 & 0.569 & 0.392 & 0.484 & 0.718 & 0.546 & 0.366 \\
        DE-SimplE$^\diamond$ & 0.526 & 0.725 & 0.592 & 0.418 & 0.513 & 0.748 & 0.578 & 0.392 \\ 
        TComplEx$^\dagger$ & 0.560 & 0.730 & 0.610 & 0.470 & 0.580 & 0.760 & 0.640 & 0.490 \\ 
        TNTComplEx$^\dagger$ & 0.560 & 0.740 & 0.610 & 0.460 & 0.600 & 0.780 & 0.650 & 0.500 \\
        TuckERT$^\ddagger$ & 0.594 & 0.731 & 0.640 & 0.518 & \underline{0.627} & 0.769 & 0.674 & 0.550 \\
        TuckERTNT$^\ddagger$ & 0.604 & \underline{0.753} & 0.655 & 0.521 & \textbf{0.638} & 0.783 & \underline{0.686} & \textbf{0.559} \\
        \midrule
        LowFER-T & 0.584 & 0.734 & 0.630 & 0.505 & 0.559 & 0.714 & 0.605 & 0.476 \\ 
        LowFER-TNT & 0.586 & 0.735 & 0.632 & 0.507 & 0.562 & 0.717 & 0.608 & 0.480 \\ 
        LowFER-CFB & \textbf{0.623} & 0.757 & \textbf{0.671} & \textbf{0.549} & \textbf{0.638} & \underline{0.791} & \textbf{0.690} & \underline{0.555} \\
        LowFER-FTP & \underline{0.617} & \textbf{0.765} & \underline{0.665} & \underline{0.537} & 0.625 & \textbf{0.792} & 0.681 & 0.534 \\
        \bottomrule
    \end{tabular}
    \caption{Time-LowFER performance on ICEWS datasets. $\diamond$ Results taken from \cite{goel2020diachronic}. $\dagger$ Results taken from \cite{lacroix2020tensor}. $\ddagger$ Results taken from \cite{shao2021tucker}.}
    \label{tab:results-full}
\end{table*}

\subsection{Chained and Factorized Bilinear Pooling}
Now we evaluate Chained Factorized Bilinear (CFB) pooling and Factorized Trilinear (FTP) pooling. 
As shown in \cref{tab:results-full}, the LowFER-CFB outperforms all baselines, including the T and TNT extensions of LowFER.
This can be attributed to the additional expressive modeling capacity of LowFER-CFB with multi-layered bilinear interactions.
CFB learns more accurate feature fusion between all three input modalities.
However, the inclusion of intermediate feature projections favors the redundancy of captured feature interactions.
Therefore, while the CFB offers improved results (w.r.t. MRR) over the FTP, it is more vulnerable to overfitting.

\subsection{Simple and Cycle-aware Time Encoding}
In this experiment, we evaluate two approaches for timestamp encoding, Simple Time Encoding (STE), which performs a bijective projection of timestamps to natural numbers, and Cyclical Time Encoding (CTE), which relies on multi-recurrent cycle-aware time decomposition (MRCA).
CTE targets specific limitations in representation learning of absolute timestamps, such as its inability to learn shared representations across different timestamps.
By introducing cyclical time components, time features to benefit from an improved sharing of parameters within individual embedding subspaces and allow for an increased generalization of short-term events.
Technically, CTE reduces the multi-collinearity of low-latent time features and allows for an improved semantic separability across individual representations. 
In CTE, we focus on high recall and therefore concentrate primarily on the Hits@10 metric in our evaluation.
Detailed results are provided in Table \ref{tab:results-cte}.
The MRCA-algorithm uses predefined time cycles, which are divided into 10 short-term (in-year) cycles and four long-term (multi-year) cycles, particularly favouring the dense distribution of time information for ICEWS datasets (see Figure \ref{fig:time-rel-hm-a}, \ref{fig:time-rel-hm-b}).
As a consequence, both ICEWS datasets show increased results (w.r.t Hits@10) with an increase of 3.4\% (ICEWS14) and 5.7\% (ICEWS05-15) for modulation-based extensions as well as 1.4\% (ICEWS14) for feature-based extensions.
\begin{figure}[t!]
    \centering
    \subfigure[][ICEWS14]{
    \label{fig:time-concentration-a}
    \includegraphics[height=1.8in]{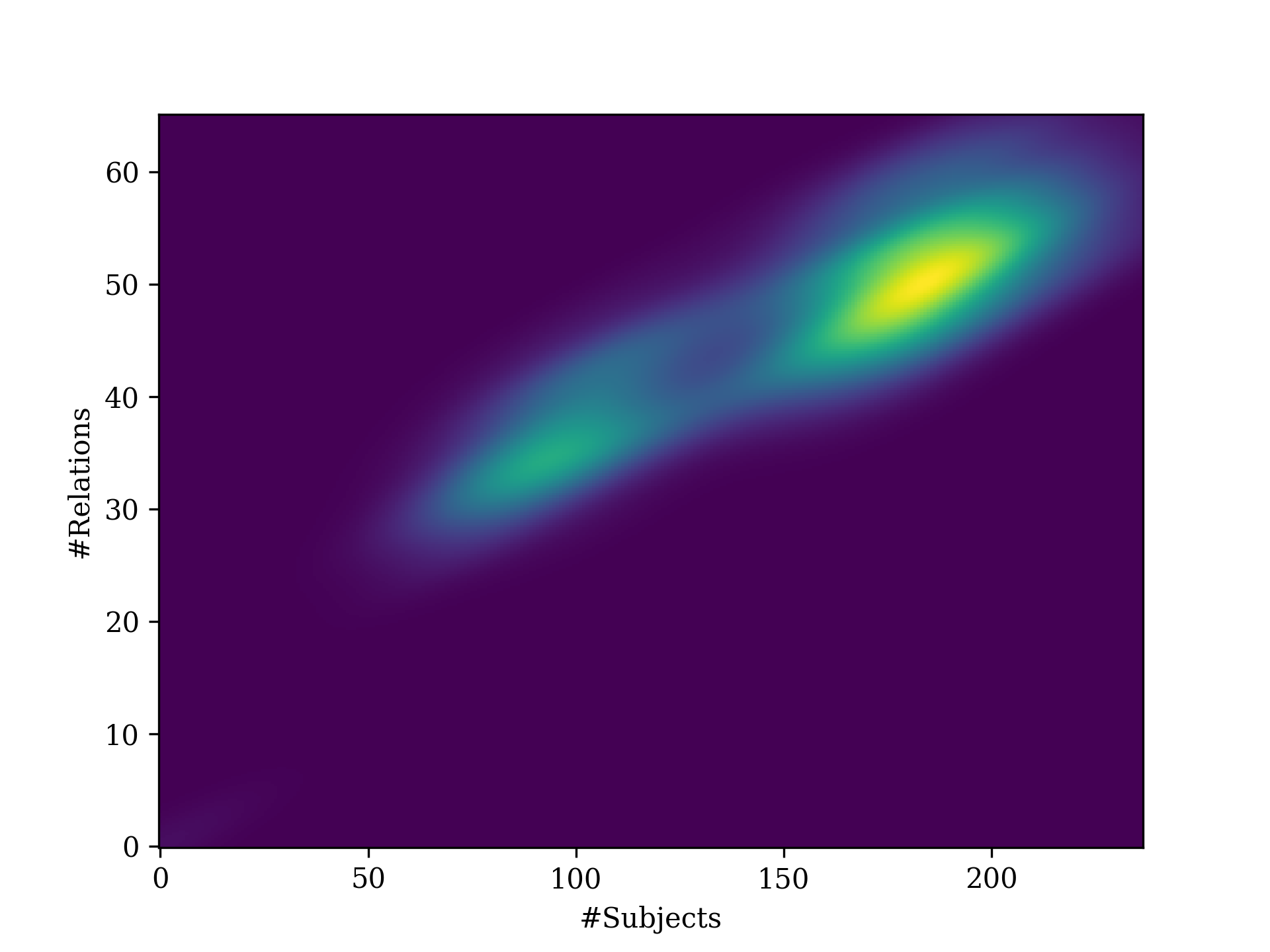}}
    \hspace{0pt}
    \subfigure[][ICEWS05-15]{
    \label{fig:time-concentration-b}
    \includegraphics[height=1.8in]{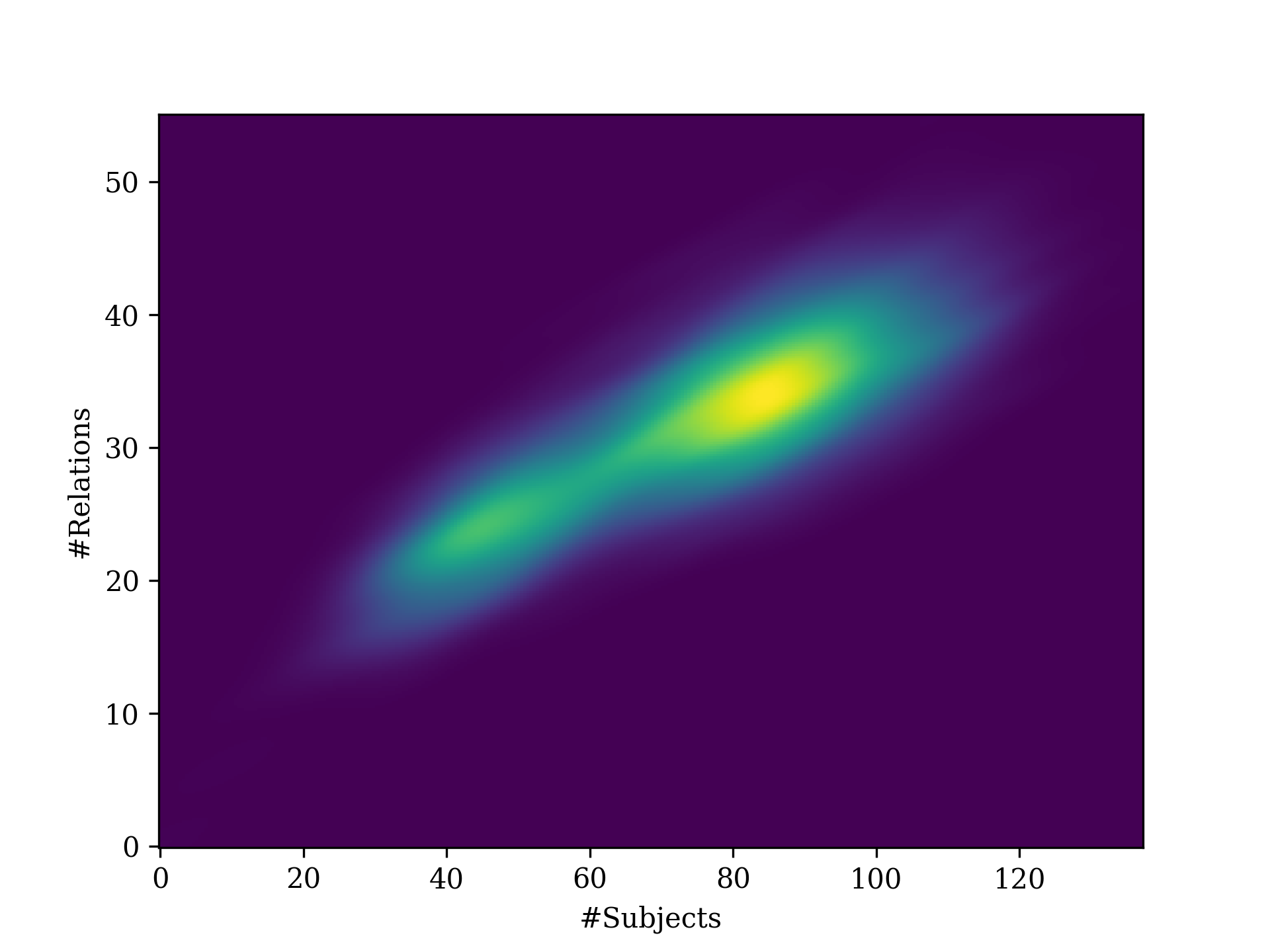}}
    \caption[Time concentration heatmap.]{Time concentration heatmap:
    \subref{fig:time-concentration-a} for ICEWS14; and
    \subref{fig:time-concentration-b} for ICEWS05-15.
    }
\label{fig:time-concentration}
\end{figure}
\begin{table*}[!t]
    \centering
    \begin{tabular}{lrcccccccc}
        \toprule
        & & \multicolumn{4}{c}{\textsc{Icews14}}  & \multicolumn{4}{c}{\textsc{Icews05-15}} \\
        \cmidrule{3-10}
        & & MRR & H@10 & H@3 & H@1 & MRR & H@10 & H@3 & H@1 \\
        \midrule
        \multirow{2}{*}{\textsc{STE}} & LowFER-T & 0.584 & 0.734 & 0.630 & 0.505 & 0.559 & 0.714 & 0.605 & 0.476 \\
        & LowFER-TNT & 0.586 & 0.735 & 0.632 & 0.507 & \textbf{0.562} & 0.717 & 0.608 & \textbf{0.480} \\
        \midrule
        \multirow{2}{*}{\textsc{CTE}} & LowFER-T & \textbf{0.600} & 0.764 & \textbf{0.654} & \textbf{0.511} & 0.556 & \textbf{0.771} & \textbf{0.621} & 0.442 \\
        & LowFER-TNT & 0.583 & \textbf{0.769} & 0.640 & 0.485 & 0.549 & 0.767 & 0.614 & 0.434 \\
        \bottomrule
    \end{tabular}
    \caption{STE and CTE results for T/TNT-extensions on ICEWS datasets.}
    \label{tab:results-cte}
\end{table*}
\begin{figure}[!t]
    \centering
    \includegraphics[width=0.8\linewidth]{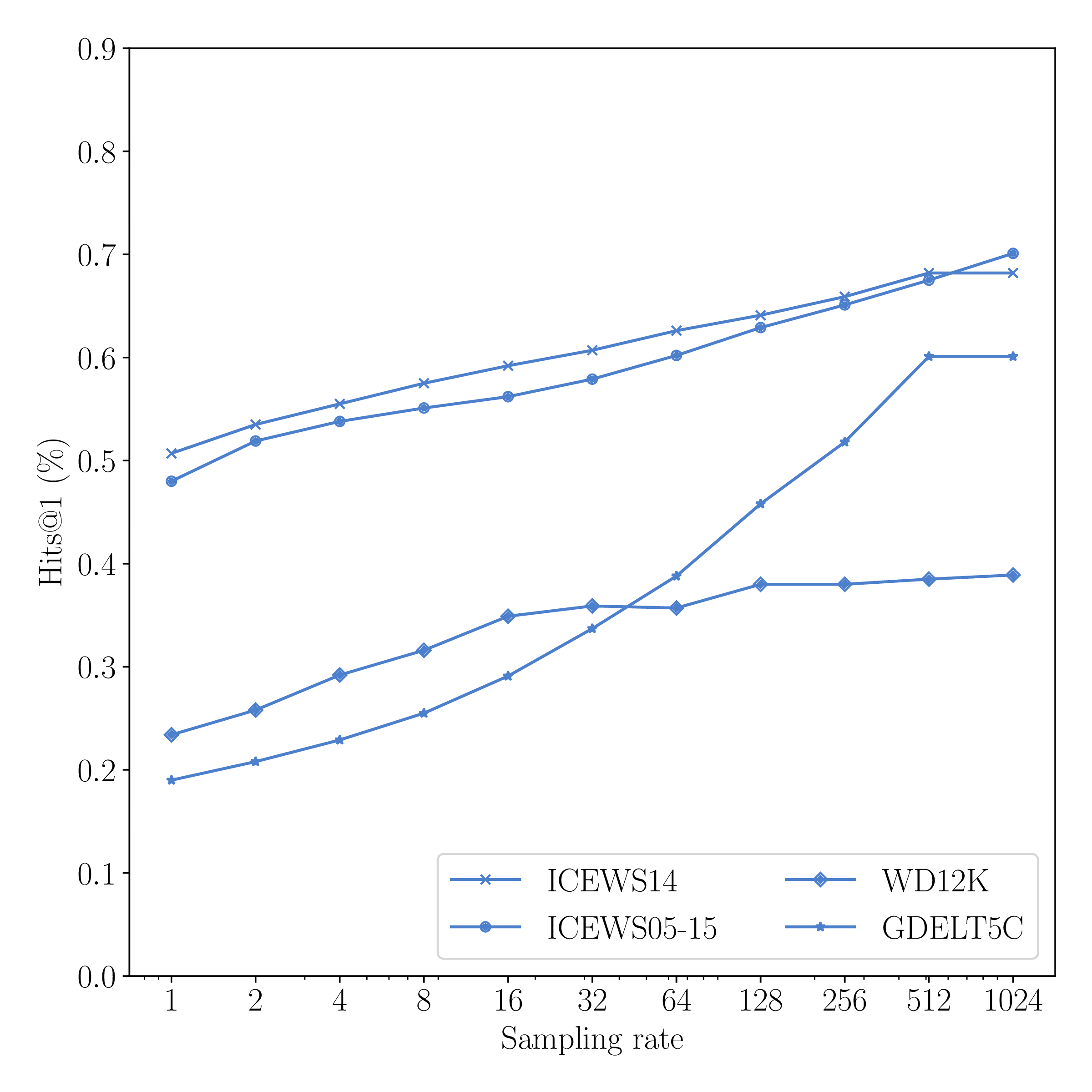}
    \caption{Influence of increasing time sampling rates.}
    \label{fig:time_ablation}
\end{figure}

\subsection{Time Sampling Rate}
In this section, we investigate the effect of time sampling in TKGC.
Following the approach of \citet{xu2020tero} and \citet{xu2020temporal}, we include a time granularity parameter, allowing to sample timestamps from a given dataset at different sampling rates.
In particular, we investigate sampling rates in the power of two $\{1.0$, $2.0$, \dots, $1024.0\}$, where $1.0$ represents the initial sampling rate, in which timestamps are discretized w.r.t to time granularity.
Sampling timestamps at lower rates cause the underlying temporal KG to aggregate facts into smaller time clusters, up to the extreme case where all facts are linked only to one timestamp.
In other words, with increasing time sampling rates, a temporal knowledge graph is synthetically transformed into a static KG.
While time sampling does not represent a viable extension for evaluating time-sensitive embedding models, it allows examining the significance of temporal facts for individual benchmark datasets.
\section{Conclusion}
In this work, we introduced Time-LowFER, a family of time-aware extensions of the bilinear factorization model LowFER.
Following existing work in temporal link prediction, we extended LowFER using time-modulated relations (TNT).
Further, noting several limitations of modulation-based extensions, we proposed two feature-based extensions of LowFER, which are based on bilinear chaining (CFB) and trilinear fusion (FTP).
In particular, we showed that the FTP represents a parameter-efficient specialization of the CFB, while CFB offers state-of-the-art results among semantic matching models for temporal link prediction.
In addition, we investigated four different approaches for time-aware extension of static embedding models and outlined, despite the increased popularity of time modulation techniques, the superiority of feature-based KGE extensions.
Furthermore, we investigated the process of time encoding in representation learning and proposed a model-agnostic method (CTE) for encoding timestamps based on multi-recurrent cycle-aware (MRCA) time decomposition.
%


\section*{Acknowledgments}
The authors would like to thank the anonymous reviewers for their helpful feedback. 
The work was partially funded by the European Union (EU) Horizon 2020 research and innovation programme through the project Precise4Q (777107) and the German Federal Ministry of Education and Research (BMBF) through the project CoRA4NLP (01IW20010).
The authors also acknowledge the cluster compute resources provided by the DFKI.
%


\bibliography{anthology,custom}
\bibliographystyle{acl_natbib}

\appendix

\section{Experimental Setup and Hyperparameters}
\label{appendix:setup}
The experiments of our work are conducted on a \textit{SLURM} computing cluster.
The virtual environments are initiated over a Unix-based system running \textit{Ubuntu} 18.04.5 (Bionic Beaver) with kernel version 5.4.0-80-generic.
Each job uses a single RTX3090 GPU (Ampere) with 24GB of shared memory and 8 CPUs.
All experiments are built using the machine learning framework \textit{PyTorch} at version 1.9.0 and NVIDIA's graphics programming interface \textit{CUDA} with toolkit version 11.1.
In addition, we performed HPT using Optuna's hyper-parameter optimization framework.
We configured the tuner to perform combined (relative/independent) sampling and used a median pruner with a warm-up threshold of 10\% and set startup trials to 10.
Further, we activated early stopping and set a maximum timeout of 24h.
For fine-tuning, we set learning rate $\in \{0.1, 0.01, 0.001, 0.0001\}$, decay rate $\in \{0.1, 0.01, 0.001, 0.0001\}$, batch size $\in \{128, 256, 512, 1024\}$ and label smoothing $\in \{0.1, 0.01, 0.001\}$.
We selected the best parameters for the final experiments and set the batch size to 1024. 

\section{ChronoKGE Framework}
\label{appendix:framework}
\textsc{ChronoKGE} 
is a unified graph embedding framework for the development of time-aware knowledge graph completion models.
It is implemented in Python and builds upon PyTorch's \cite{paszke2019pytorch}.
%
Our framework focuses on time-sensitive representation learning tasks for temporal and event knowledge graphs and offers an easy-to-use and flexible library with various time-focused functionalities, including time-specific sampling and encoding routines.
\textsc{ChronoKGE} supports multiple temporal knowledge graphs with diverse graph schemas and offers a dynamic interface for adding new knowledge graphs.
Similarly, our framework provides an easy interface to add new or extend existing learning models.
Therefore, several generic embedding models are already available within the \texttt{model.kge} package, which extend PyTorch's default \texttt{nn.Module} by integrating commonly required methods in knowledge representation learning.
In addition, it provides a customizable and flexible package for defining experimental jobs as well as additional modules for training.
To support hyper-parameter optimization, we integrated a parameter tuning system which is based on the \textit{Optuna}\footnote{\url{https://optuna.org}} framework.
The tuning system is part of the integrated \texttt{chrono\_kge.tuner} package and allows for an automatic search of optimal hyper-parameters.

\section{Limitations} \label{appendix:limitations}
In its current form, our proposed methods can overfit since they lack commonly used regularization schemes, such as time-smoothness and nuclear 3-norm \cite{lacroix2020tensor,shao2021tucker}.
However, extending our work with regularization schemes is straightforward. 
\begin{figure}[!t]
    \centering
    \includegraphics[width=0.48\textwidth]{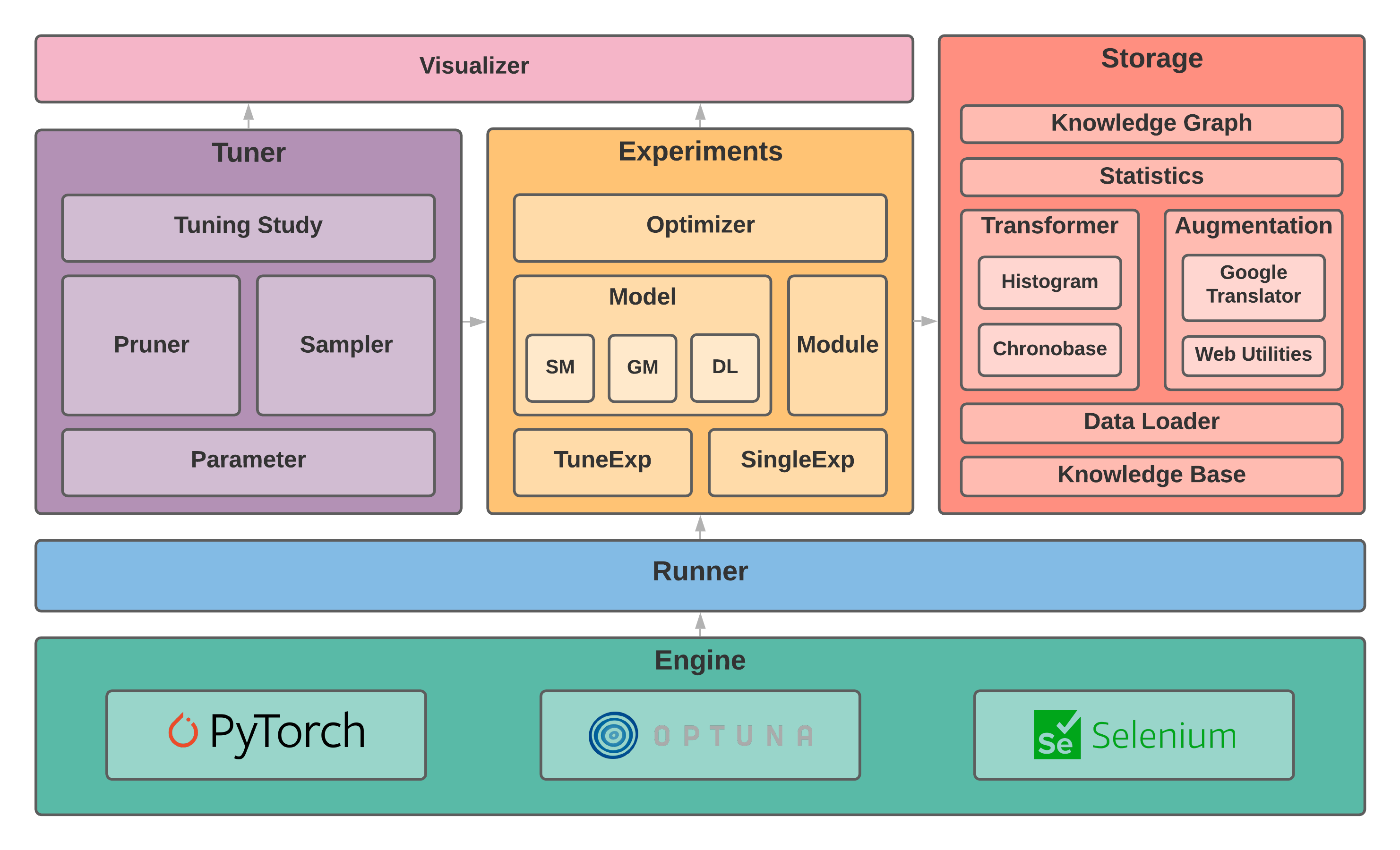}
    \caption{Overview of the ChronoKGE framework.}
    \label{fig:framework}
\end{figure}
In terms of the design choices, LowFER-CFB offers a more expressive representation. However, it is computationally more expensive (7.85 sec/epoch for LowFER-CFB compared to 4.19 sec/epoch for LowFER-TNT on ICEWS-14) and MRCA, despite offering a generalized time representation, has limited performance gains and further adds a computational footprint (2.19 additional seconds per epoch on ICEWS-14 for LowFER-TNT).




\end{document}